\documentclass[journal]{IEEEtran}

\ifCLASSINFOpdf
\else
\fi
\usepackage{multirow}
\usepackage{threeparttable}
\usepackage{booktabs}
\usepackage{cite}
\usepackage{graphicx}  
\usepackage{epstopdf}  
\usepackage{multirow}
\usepackage{array}  
\usepackage{color}
\usepackage{colortbl}
\usepackage{amsmath}
\usepackage{tabularx}
\usepackage[table]{xcolor}
\usepackage{amssymb}
\usepackage{mathrsfs}  
\usepackage{amsfonts}  
\usepackage[switch]{lineno}
\usepackage{algorithm}
\usepackage{algpseudocode}
\usepackage{pifont}     
\usepackage{wasysym}    
\usepackage{url}

\makeatletter
\DeclareRobustCommand\onedot{\futurelet\@let@token\@onedot}
\def\@onedot{\ifx\@let@token.\else.\null\fi\xspace}

\makeatother

\begin{document}
%
\title{Improving Large Vision-Language Models' Understanding for Field Data}
%
%

\author{Xiaomei Zhang, Hanyu Zheng, Xiangyu Zhu,~\IEEEmembership{Senior Member,~IEEE}, Jinghuan Wei, Junhong Zou, Zhen Lei,~\IEEEmembership{Senior Member,~IEEE} and Zhaoxiang Zhang

\thanks{The authors are with the State Key Laboratory of Multimodal Artificial Intelligence Systems, Institute of Automation, Chinese Academy of Sciences (Beijing 100190).}
\thanks{Manuscript received Jul. xx, 2025.} 
}

\maketitle

\begin{abstract}
Large Vision-Language Models (LVLMs) have shown impressive capabilities across a range of tasks that integrate visual and textual understanding, such as image captioning and visual question answering. These models are trained on large-scale image and video datasets paired with text, enabling them to bridge visual perception and natural language processing. However, their application to scientific domains, especially in interpreting complex field data commonly used in the natural sciences, remains underexplored. In this work, we introduce FieldLVLM, a novel framework designed to improve large vision-language models' understanding for field data. FieldLVLM consists of two main components: a field-aware language generation strategy and a data-compressed multimodal model tuning. The field-aware language generation strategy leverages a special-purpose machine learning pipeline to extract key physical features from field data, such as flow classification, Reynolds number, and vortex patterns. This information is then converted into structured textual descriptions that serve as a dataset. The data-compressed multimodal model tuning focuses on LVLMs with these generated datasets, using a data compression strategy to reduce the complexity of field inputs and retain only the most informative values. This ensures compatibility with the model’s language decoder and guides its learning more effectively. Experimental results on newly proposed benchmark datasets demonstrate that FieldLVLM significantly outperforms existing methods in tasks involving scientific field data. Our findings suggest that this approach opens up new possibilities for applying large vision-language models to scientific research, helping bridge the gap between large models and domain-specific discovery.
\end{abstract}


\begin{figure*}[ht]
    \centering
    \includegraphics[width=1\textwidth]{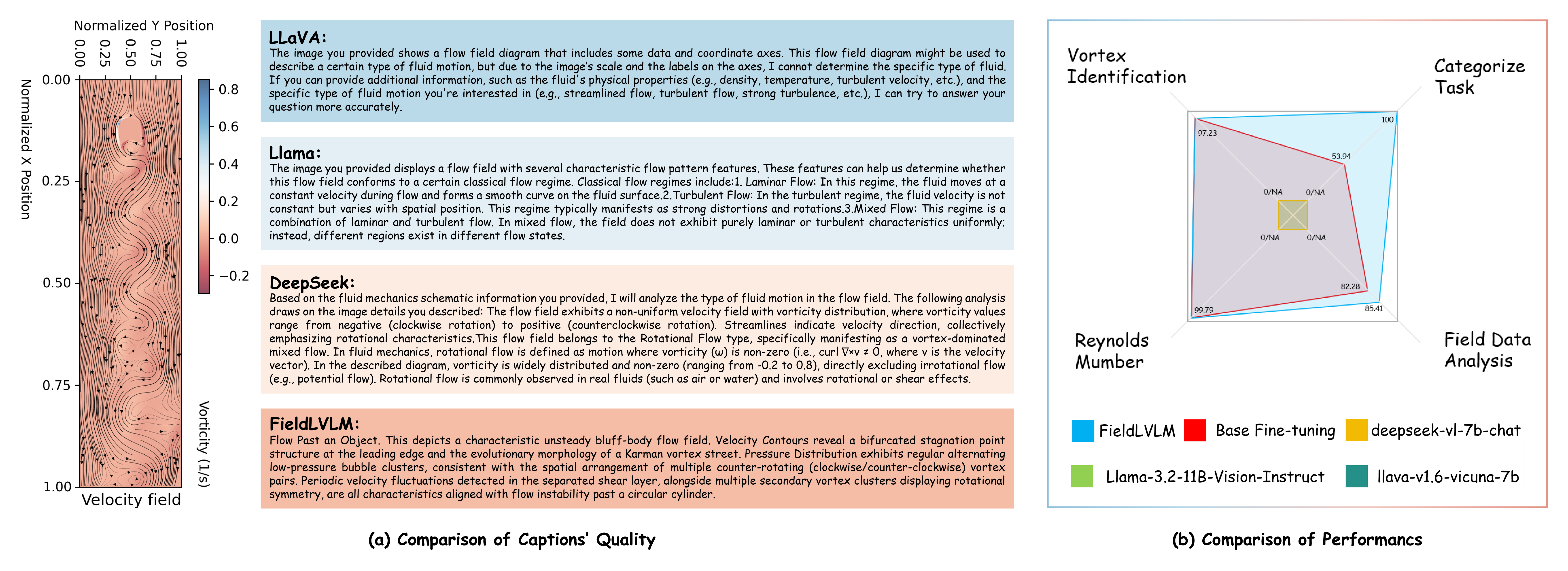}
    \caption{(a) Quantitative comparison of vision-language responses in field data across different methods. (b) The performance of key evaluation metrics, including flow categorization, Reynolds number calculation, vortex identification, and comprehensive field data interpretation.}
    \label{fig:example}
\end{figure*}

\section{Introduction}
Vision and language are two fundamental modalities through which humans perceive and interpret the world. While vision provides spatial and structural understanding, language conveys contextual and abstract information. Recent advances in large vision-language models (LVLMs) have enabled the integration of visual inputs (e.g., images, videos) with natural language, allowing these models to capture both spatial relationships and semantic meaning. This fusion significantly enhances their representational power and enables comprehensive scene understanding. LVLMs have been shown to learn a substantial amount of in-depth knowledge from data. Pre-trained LVLMs have demonstrated impressive performance across a range of open-world visual tasks, including image classification~\cite{radford2021learning,yuan2021florence}, object detection~\cite{li2022grounded,zhong2022regionclip}, and semantic segmentation~\cite{zou2023generalized,li2022language}. However, LVLMs performed unsatisfactorily on field scientific data. However, their performance remains limited when applied to scientific field data. This limitation arises primarily from two challenges. First, the availability of high-quality, large-scale scientific datasets suitable for training LVLMs is limited. Second, field data often varies in length and complexity, and may exceed the maximum input token limits of current models, making effective encoding and reasoning difficult.

In this paper, we propose a novel framework for enhancing large vision-language models' understanding of scientific field data. Our approach integrates two key components: a field-aware language generation strategy and a data-compressed multimodal model tuning. The proposed field-aware language generation strategy combines the advantages of the high accuracy of the special-purpose models and the consistency of the large language models, as our data source. While the domain-special models excel in accuracy for specialized tasks, but lack generalization and consistency across different data types. In contrast, the large language models provide broad and consistent interpretations but may underperform in precision for domain-specific analysis. Specifically, given field data, the field-aware language generation strategy uses domain-special models to generate flow classification, Reynolds number, and vortex detection. Then, the results and original field data are sent into a large language model to produce a consistent field language representation for downstream learning and model tuning.

Building on the recently proposed Qwen2.5-VL~\cite{bai2025qwen2}, we develop a data-compressed multimodal model tailored to the unique characteristics of field data. To accommodate the input constraints of large language models, we incorporate VQGAN~\cite{esser2021taming} to compress field data. Specifically, we first map the velocity and pressure fields onto the three channels of an RGB image to produce a 256×256 representation of the original field. This image is then encoded by VQGAN into 256 discrete tokens, which are passed to the language decoder. In parallel, we extract several representative physical values from the original data to guide the model’s learning, ensuring that key quantitative features are preserved. To enrich the semantic structure of the input, we also convert the generated textual descriptions of the field into image representation, which is sent into the image decoder. The compressed tokens, representative key values, and semantic image representations are then fed into the language-augmented multimodal model. Experimental results on our proposed benchmark datasets demonstrate that our method outperforms existing state-of-the-art approaches in field data understanding tasks, achieving superior performance across multiple evaluation metrics.
In particular, our paper makes the following contributions:


\begin{itemize}
  \item[1.] A novel framework (FieldLVLM) that bridges vision-language modeling with scientific field data understanding.
  \item[2.] We present a data reformation pipeline, a field-aware language generation strategy, which integrates high-accuracy special-purpose models with large language models for consistent and interpretable field descriptions.
  \item[3.] We develop a data-compressed multimodal model tuning by compressing the original data to meet the requirements of the language decoder for token limitations. Meanwhile, we select representative key values to guide the learning of the network.
  \item[4.] Extensive experiments demonstrating that our method achieves excellent results on challenging scientific datasets.
\end{itemize}

\section{Related Work}
\subsection{Large Language Models}
In recent years, advancements in data availability and computational resources have significantly propelled the growth of large language models. Early models such as BERT~\cite{devlin2019bert} and T5~\cite{raffel2020exploring}, based on an encoder-decoder structure, along with decoder-focused architectures like GPT~\cite{radford2018improving}, utilized the Transformer framework to achieve notable success across a range of natural language processing tasks. The breakthrough of GPT3~\cite{brown2020language} helped establish decoder-only architectures, which generate outputs through auto-regressive methods. Building on this foundation, models such as PaLM~\cite{chowdhery2023palm} scaled up both in terms of parameters and training data, while others like InstructGPT~\cite{ouyang2022training} and ChatGPT~\cite{biswas2023role} incorporated techniques like fine-tuning and reinforcement learning to enhance dialogue capabilities.

\subsection{Large Visual-Language Models}

With the rapid advancement of large language models (LLMs), an emerging group of researchers has begun to focus on integrating visual information into these models. Foundational efforts in this direction come from the field of vision-language learning, particularly in modality alignment~\cite{jia2021scaling,radford2021learning}. One prominent example is CLIP~\cite{radford2021learning}, which demonstrates how contrastive learning on large-scale image-text datasets can effectively align visual and textual representations. Models such as LLaVA~\cite{liu2023visual} and
InstructBLIP~\cite{2023InstructBLIP} further pushes the boundaries by refining instruction tuning to better interpret complex queries. Despite these advancements, current field data still suffers from a shortage of high-quality image-text pairs, limiting the applicability of large vision-language models.

\subsection{Scientific Discovery}
Scientific discovery has traditionally depended on methods like hypothesis-driven experimentation, statistical analysis, and specialized simulations. Early computational tools, including symbolic reasoning systems and expert-based frameworks~\cite{feigenbaum1977art}, played a foundational role in augmenting scientific inquiry. Yet, these conventional techniques often demand significant manual input and deep domain knowledge, making them difficult to scale or adapt across fields. Recently, the rise of data-centric methodologies, particularly those employing deep learning, has reshaped the process of discovery. Large language models (LLMs) and general-purpose foundation models have shown impressive performance in areas such as automated hypothesis formulation~\cite{zheng2025large},
predicting material properties~\cite{tshitoyan2019unsupervised}, and modeling protein structures~\cite{jumper2021highly}. By leveraging vast datasets of scientific texts and experimental results, these models can uncover hidden patterns and offer insights that might surpass human intuition. Despite their potential, ongoing research is still addressing key issues, including interpretability, generalization to specific scientific domains, and the accuracy of generated content.

\begin{figure*}[t]
 \begin{center}
 \includegraphics[width=0.98\linewidth]{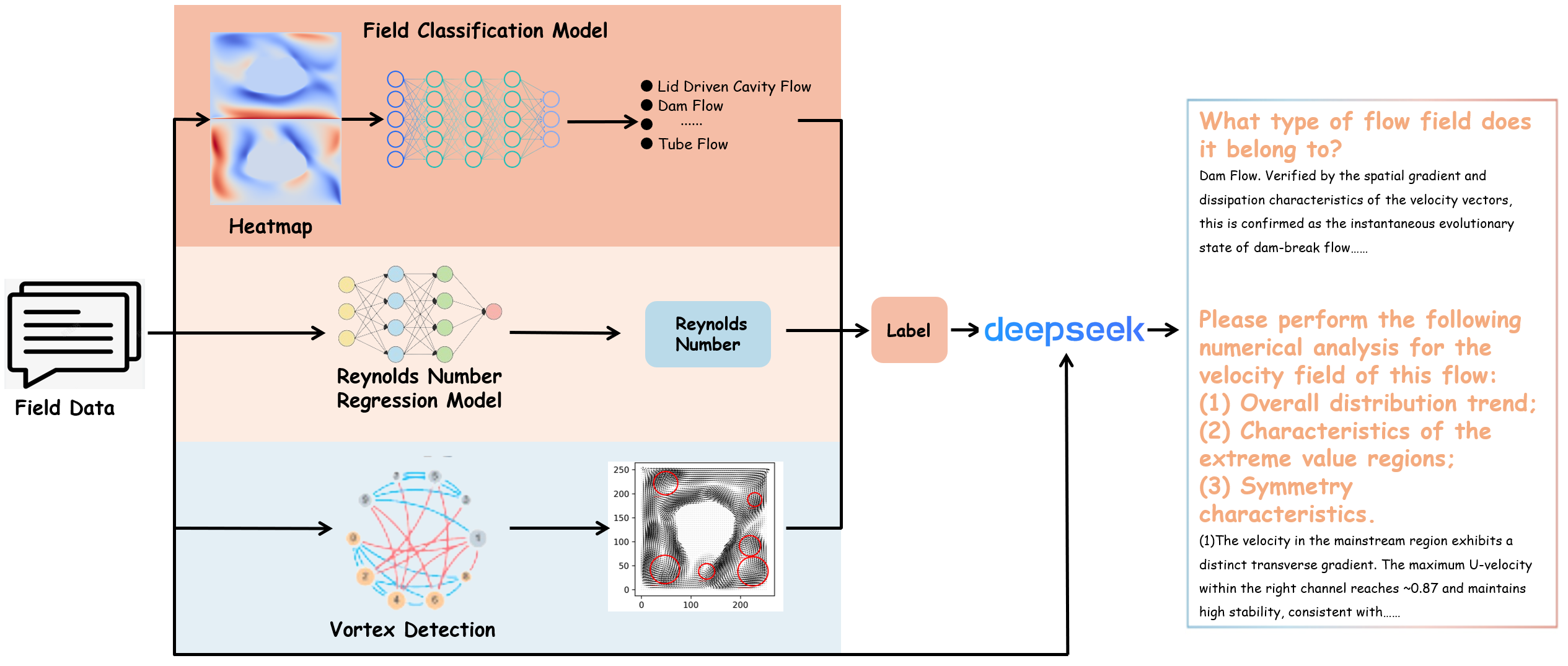}
 \caption{The pipeline of field-aware language data generation strategy integrating special-purpose machine learning models for field classification, Reynolds number estimation and vortex detection.}
 \label{fig:1}
 \end{center}
\end{figure*}

\begin{figure}[t]
 \begin{center}
 \includegraphics[width=0.98\linewidth]{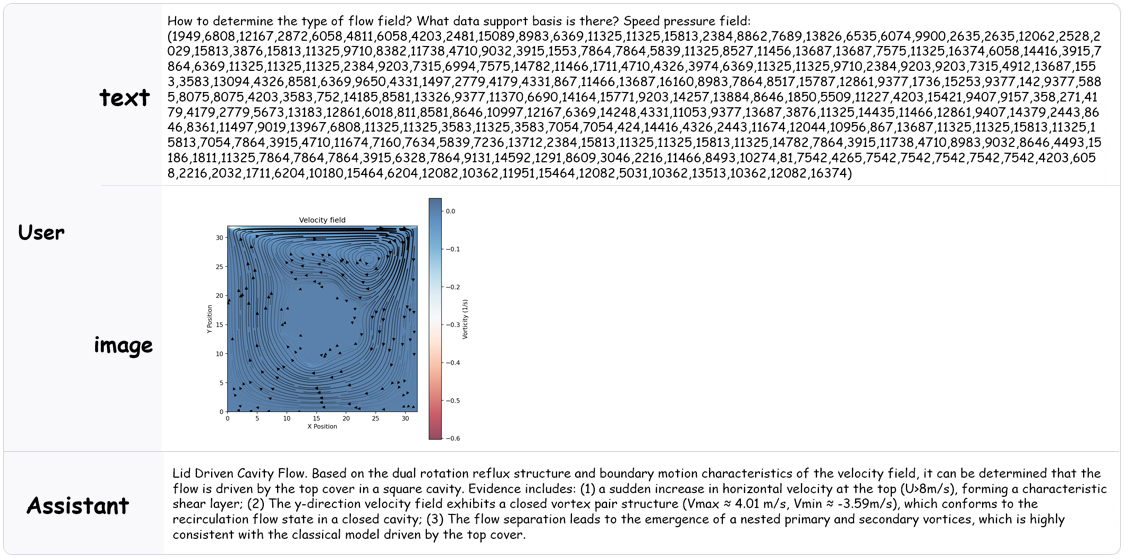}
 \caption{Illustrative examples of generated field language representations, showing structured text outputs for flow field analysis derived from velocity and pressure data.}
 \label{fig:2}
 \end{center}
\end{figure}

\subsection{Image-text Data Enhancement}
In the field of vision-language learning, various efforts~\cite{fan2023improving,gadre2023datacomp,lai2023scarcity,nguyen2023improving} have been made to improve the quality of image-text pair captions. For instance, LaCLIP ~\cite{fan2023improving} utilizes large language models to refine original captions, though its performance is often compromised by hallucinations stemming from limited visual context and the poor quality of raw annotations. Other studies~\cite{gadre2023datacomp,lai2023scarcity} have explored strategies for combining and filtering both original and synthetic captions to boost CLIP performance. A more recent approach, VeCLIP~\cite{nguyen2023improving}, employs LLMs to merge insights from both caption sources. However, due to the low fidelity of synthetic data, the resulting captions integrate only minimal visual content. To date, LLaVA~\cite{liu2023visual} stands out in the multimodal model domain by feeding short, human-written captions and bounding boxes into the GPT-4 language model, enabling it to simulate visual perception before generating richer descriptions. Despite this, the method remains heavily reliant on labor-intensive human annotations and lacks true visual input, often leading to overly detailed descriptions of main objects—including those in marginal areas—guided solely by bounding box cues, thereby increasing the risk of hallucinations. In contrast, our work adopts GPT-4 Vision, a state-of-the-art multimodal model capable of generating rich, accurate captions directly from crafted prompts and image inputs, thereby significantly enhancing the integration of visual knowledge.

\begin{figure*}[t]
 \begin{center}
 \includegraphics[width=1\linewidth]{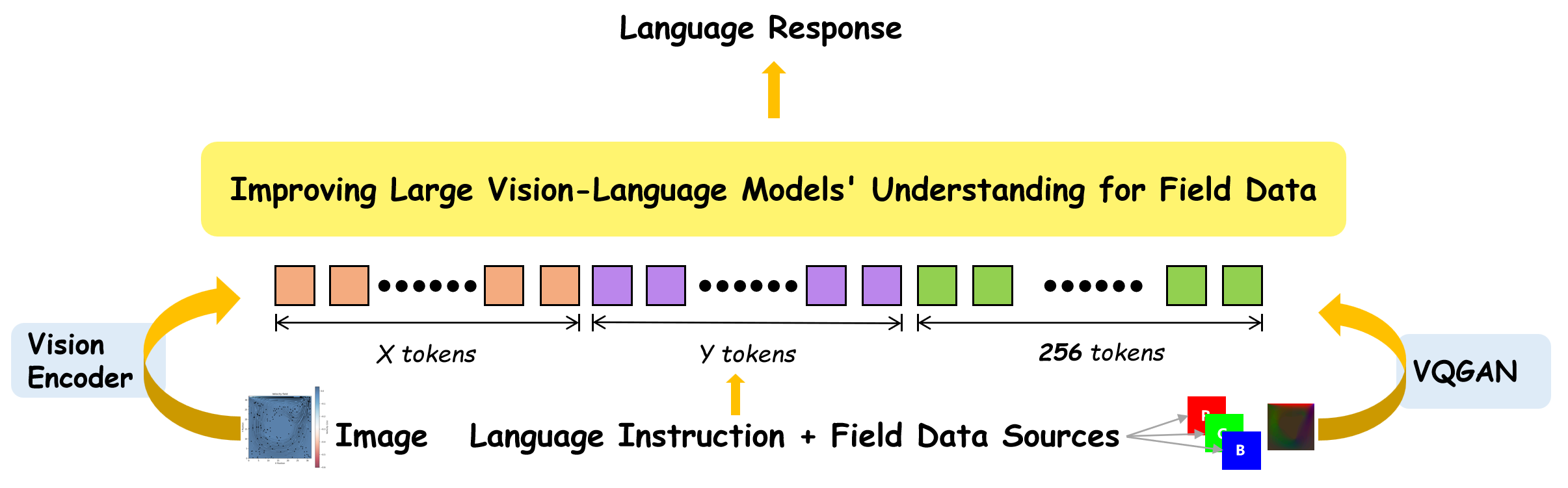}
 \caption{Input-output architecture of the data-compressed multimodal model featuring VQGAN-based token compression, key value selection, and image representation conversion for enhanced field data semantic analysis. }
 \label{fig:2}
 \end{center}
\end{figure*}

\section{Method}\label{sec:method}

\subsection{Field-aware Language Generation Strategy}\label{sec:overall}
While the availability of large-scale multimodal datasets—such as image-text pairs from CC~\cite{changpinyo2021conceptual} and LAION~\cite{schuhmann2022laion}—has significantly advanced vision-language research, the same progress has not extended to scientific domains. Multimodal scientific datasets remain scarce, primarily due to the difficulty in accessing such data and the substantial expertise and time required for accurate annotation. To address this limitation, we draw inspiration from the DeepSeek model~\cite{zhu2024deepseek}, which has demonstrated strong performance in automated text annotation. Leveraging its capabilities, we propose using DeepSeek to generate field-specific textual descriptions based on existing field data. This computerized approach enables scalable field language collection, reducing reliance on manual annotation while improving the quality and consistency of the resulting multimodal dataset.

Given a field of data $X$, it is natural to create a set of questions $X_q$ with the intent to describe the field content. We leverage Deepseek to make such a list of questions. Therefore, a simple way to expand a field of data to its data-language version is Human: $X$, $X_q$ $<STOP>$ Assistant: $X_l$, where $X_l$ represents the generated language description. Though cheap to construct, this simple expanded version lacks diversity and in-depth reasoning.

To address the limitations of shallow reasoning in simple prompt-based generation, we incorporate domain-specific models as strong teachers to enrich the language representation of field data. Specifically, we leverage high-accuracy models for tasks such as classification~\cite{he2016deep} and detection~\cite{zhang2020bridging}, which operate on visual inputs. To enable this, we first convert the raw field data into image representations. Guided by expert knowledge, we begin by categorizing flow fields based on environmental and geometric characteristics—such as distinguishing between cavity-driven flows and external flows. For instance, in the case of a top-cover-driven cavity flow, we identify its defining features to support more targeted analysis. Next, we classify the flow regime by computing the Reynolds number. This regression model provides essential insight into the underlying dynamics. In addition, we account for rare but critical phenomena—such as sudden flow instabilities or anomalies—that can emerge as low-probability events. These are detected using specialized models, allowing for early identification and proactive mitigation strategies.




\subsection{Data-Compressed Multimodal Model Tuning}\label{sec:semantic}

\subsubsection{Data}
Large language models have demonstrated the ability to acquire substantial domain knowledge from data. However, when applied to scientific field data, they face two key limitations. First, plain textual representations of field data often lack explicit semantic structure, making it difficult for the model to extract meaningful patterns or relationships. As a result, important physical characteristics may be overlooked or poorly understood. Second, these models are constrained by a fixed maximum input length. Field data—often high-dimensional and detailed—can easily exceed this token limit, leading to truncation or loss of critical information.

Images are more conducive to the discovery of semantic information than text. Therefore, we convert the textual representation of field data into an image representation. Then, the image representations of the field data are sent into the image encoder of the multimodal large model.


To address the challenge of excessive token length in field science data (e.g., velocity-pressure matrices exceeding 65,536 tokens), we implement a two-stage compression pipeline that reduces input dimensionality by 99.6\% while preserving critical physical features. In the first stage, the raw 256×256 scalar fields of horizontal velocity (u), vertical velocity (v), and pressure (p) are linearly normalized and mapped to a 3-channel RGB image via $R \propto u_{\text{norm}}$, $G \propto v_{\text{norm}}$, $B \propto p_{\text{norm}}$, respectively. This RGB representation is then encoded into 256 discrete tokens using a pre-trained VQGAN model, effectively compressing the original 65,536-character sequence into a compact latent representation compatible with Qwen2.5-VL’s token limit.

The local key data in the data is of guiding significance for analyzing the entire data. Therefore, based on? The principle is to select a key piece of data for focused analysis to enhance the effectiveness of the overall data analysis.

\subsubsection{Training}

We conduct parameter-efficient fine-tuning using Low-Rank Adaptation~\cite{hu2022lora} (LoRA) within the LLaMA-Factory framework, focusing on adapting the Qwen2.5-VL-7B model to field data tasks. The architecture maintains two critical design principles: Frozen Visual Encoder, the CLIP-ViT backbone remains entirely locked during training to preserve pre-trained visual representations and prevent catastrophic forgetting; Selective Parameter Update, only LoRA adapters (rank=32, scaling factor=128) and the multimodal projector are updated, reducing trainable parameters by 98.7\% compared to full fine-tuning.

The model receives multimodal input comprising two components: (1) a textual prompt that poses an analytical question, paired with a compressed token sequence representing the velocity-pressure field; and (2) a heatmap visualization of the flow field, which provides additional visual context. The model is trained to generate structured outputs aligned with one of four predefined analytical categories. For example, when tasked with vortex analysis, the output includes detailed parameters such as vortex location, size, circulation strength, and rotation direction.

\begin{table*}[t]
\begin{center}
\caption{Comparative evaluation with some vision-language models, including DeepSeek-VL-7B~\cite{zhu2024deepseek}, LLaVA-v1.6~\cite{liu2023visual}, and Llama-3.2~\cite{Llama-3.2} in categorize task, Reynolds number calculation, vortex identification, and comprehensive field data analysis.} \label{t1}
\resizebox{0.88\textwidth} {!} {
\begin{threeparttable}
    \begin{tabular}{ c|c|c|c|c }
    \toprule
    Method &Categorize Task&Reynolds Number&Vortex Identification&Field Data Analysis \cr
    \midrule
    \midrule
    DeepSeek-VL-7b-chat~\cite{zhu2024deepseek}  &0/NA &0/NA &0/NA &0/NA \cr
    LLaVA-v1.6-vicuna-7b~\cite{liu2023visual} &0/NA &0/NA &0/NA &0/NA \cr
    Llama-3.2-11B-Vision-Instruct~\cite{Llama-3.2} &0/NA &0/NA &0/NA &0/NA \cr
    FieldLVLM (ours) &100 &99.79 &97.23 &85.41 \cr
    \bottomrule
    \end{tabular}
\end{threeparttable}
}
\end{center}
\end{table*}

\begin{table}[t]
\begin{center}
\caption{Ablation analysis of data compression impact on vortex identification accuracy in FieldLVLM framework: evaluating Qwen2.5-VL-7B baseline, base fine-tuning, and enhanced rerformance with VQGAN-based token compression.} \label{t2}
\resizebox{0.48\textwidth} {!} {
\begin{threeparttable}
    \begin{tabular}{ c|c }
    \toprule
    Method &Vortex Identification \cr
    \midrule
    \midrule
    Qwen2.5-VL-7B (Baseline) \cite{bai2025qwen2}  &0/NA
\cr
    Base Fine-tuning & 82.28\cr
    + Compress data  &85.41\cr
    \bottomrule
    \end{tabular}
\end{threeparttable}
}
\end{center}
\end{table}

\begin{table}[t]
\begin{center}
\caption{Ablation study on key data selection for field data analysis in FieldLVLM: performance comparison of Qwen2.5-VL-7B baseline, base fine-tuning, and optimized accuracy with representative value guidance addressing semantic structure challenges.} \label{t3}
\resizebox{0.48\textwidth} {!} {
\begin{threeparttable}
    \begin{tabular}{ c|c }
    \toprule
    Method &Field Data Analysis \cr
    \midrule
    \midrule
    Qwen2.5-VL-7B (Baseline) \cite{bai2025qwen2}  &0/NA
 \cr
    Base Fine-tuning &53.94 \cr
    + Key data  &100 \cr
    \bottomrule
    \end{tabular}
\end{threeparttable}
}
\end{center}
\end{table}

\section{Experiments}\label{experiment}
\subsection{Implementation Details and Dataset Statistics}

\textbf{Implementation Details.} We evaluate FieldLVLM using the Qwen2.5-VL-7B architecture as the base model. The training is conducted over 4 epochs with a learning rate of 5e-5, a maximum gradient norm of 1.0, and a context length of 4096 tokens. Each device processes a batch of 4 samples, with 8 gradient accumulation steps, resulting in an effective batch size of 32. A validation split of 10\% is used, and learning rate warm-up is applied during the first 100 steps.

\textbf{Dataset Statistics.}
Our experiments are based on extended versions of FlowBench~\cite{tali2024flowbench} and CFDBench~\cite{luo2023cfdbench}, where each sample is paired with structured language descriptions generated by our field-aware strategy.

\subsection{Evaluation Benchmark and Metrics}
We designed four benchmark tasks to evaluate scientific field data understanding: Flow Categorization, Reynolds Number Estimation, Vortex Identification, and Comprehensive Field Analysis. Each task targets a specific analytical capability required for interpreting velocity-pressure fields, and is benchmarked using over 70,000 samples from the aforementioned datasets.

\textbf{Categorize Task} focuses on classifying flow types based on morphological and dynamic characteristics of velocity-pressure fields, employing Accuracy (\%) defined as the ratio of correctly identified flow types (e.g., lid-driven cavity flow vs. flow past a bluff body) to ground-truth annotations. Fluid states exhibit distinct spatial signatures detectable through vision-language fusion – such as Kármán vortex streets in bluff-body flows or shear layers in lid-driven cavities – where high accuracy demonstrates FieldLVLM’s mapping of visual patterns to canonical flow categories, critical for preliminary diagnostics.

\textbf{Reynolds Number} calculation computes the dimensionless parameter ($Re = \frac{\rho U L}{\mu}$) from field data, using Accuracy (\%) with predictions deemed correct at $\leq10\%$ relative error versus ground truth. This combines numerical precision with thermodynamics-based validation to mitigate hallucination risks prevalent in prior LVLMs, ensuring physically plausible outputs.

\textbf{Vortex Identification} detects coherent structures while quantifying position, radius, and rotation direction. We define Accuracy (\%) as successful when: (i) vortex center coordinates deviate $\leq25\%$ of flow domain size from annotations, and (ii) rotation direction matches ground truth. This validates FieldLVLM’s geometric feature extraction from complex topologies, addressing conventional LVLMs’ spatial reasoning limitations.

\textbf{Field Data Analysis} determines maximum velocity ($U_{\max}$) and interprets field semantics via natural language Q\&A. Metrics include Accuracy (\%) for $U_{\max}$ localization ($\leq10\%$ error tolerance) and Cross-Entropy Loss measuring semantic consistency in velocity-pressure analysis across flow regimes. $U_{\max}$ evaluation tests fine-grained regression capability, while cross-entropy quantifies contextual reasoning essential for deriving physics-compliant insights from multimodal inputs.

\begin{figure*}[t]
    \centering
    \includegraphics[width=0.85\textwidth]{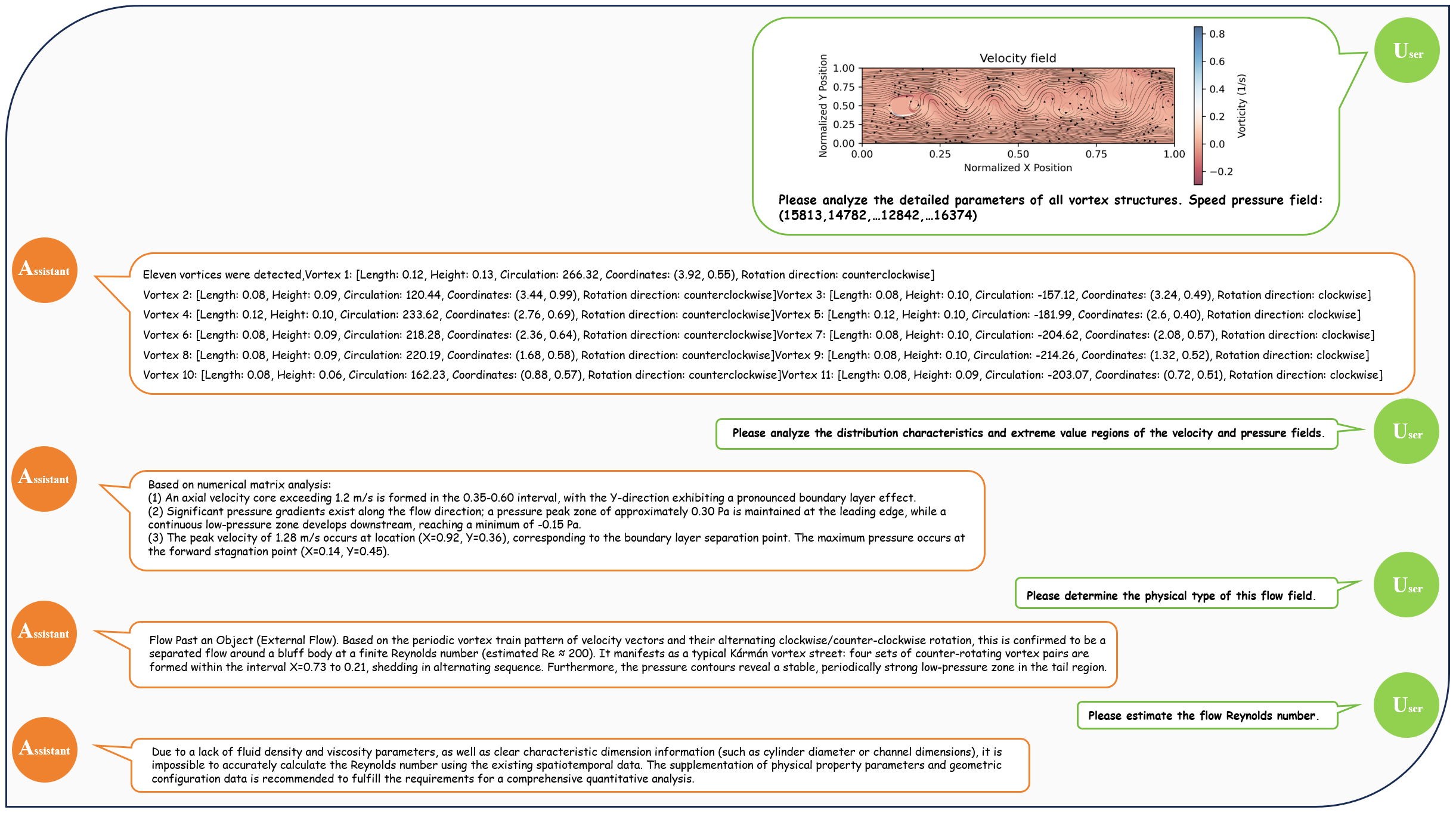}
    \caption{Q\&A analysis on vortex shedding dynamics and pressure distribution in flow past a bluff body highlighting kármán vortex street characteristics and three-stage flow structure.}
    \label{fig:5}
\end{figure*}
\begin{figure*}[t]
    \centering
    \includegraphics[width=0.85\textwidth]{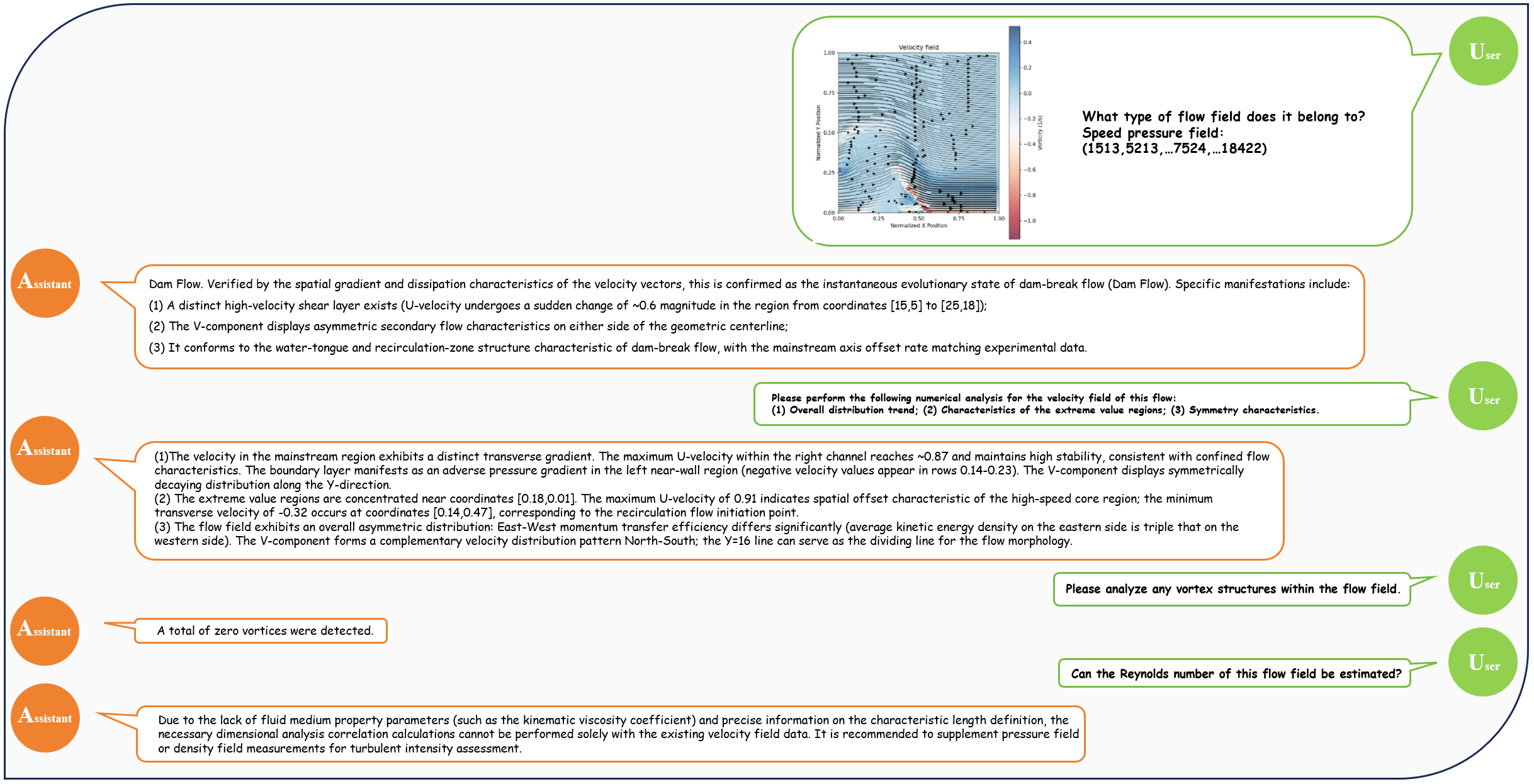}
    \caption{Q\&A session on velocity distribution and vorticity analysis for dam break flow field characterization demonstrating shear layer dynamics and critical flow regimes.}
    \label{fig:6}
\end{figure*}

\subsection{Results}
We systematically evaluated the proposed FieldLVLM framework against state-of-the-art baselines, such as DeepSeek-vl-7b-chat~\cite{zhu2024deepseek}, LLaVA-v1.6-vicuna-7b~\cite{liu2023visual}, Llama-3.2-11B-Vision-Instruct~\cite{Llama-3.2}. We make comparisons on four standardized tasks: Categorization, Reynolds Number calculation, Vortex Identification, and Field Data Analysis.
As demonstrated in Table~\ref{t1}, our Improving Large Vision-Language Models' Understanding for Field Data (FieldLVLM) significantly outperformed existing general-purpose models in scientific field data processing. FieldLVLM achieved 99.79\% accuracy in Reynolds Number calculations, while baseline models (deepseek-vl-7b-chat, Ilava-v1.6-vicuna-7b, Llama-3.2-v1B-Vision-Instruct) registered 0/NA due to architectural incompatibilities with scientific data structures. Similarly, FieldLVLM attained 97.23\% in Vortex Identification and 85.41\% in Field Data Analysis, outperforming all comparative models, which scored 0/NA across tasks. These results underscore the limitations of conventional architectures in handling long-sequence scientific data with semantic irregularities, while confirming our method enhances robustness in scientific discovery tasks.

\begin{table*}[t]
\centering
\caption{Comparative analysis of flow field classification performance across large vision-language models in lid-driven cavity flow scenario.}\label{t4}

\begin{tabularx}{\textwidth}{@{}>{\bfseries}l X@{}}
\toprule
\multicolumn{2}{c}{\textbf{Visual input example, Extreme Ironing:}} \\
\midrule
\multicolumn{2}{c}{\includegraphics[width=0.4\linewidth]{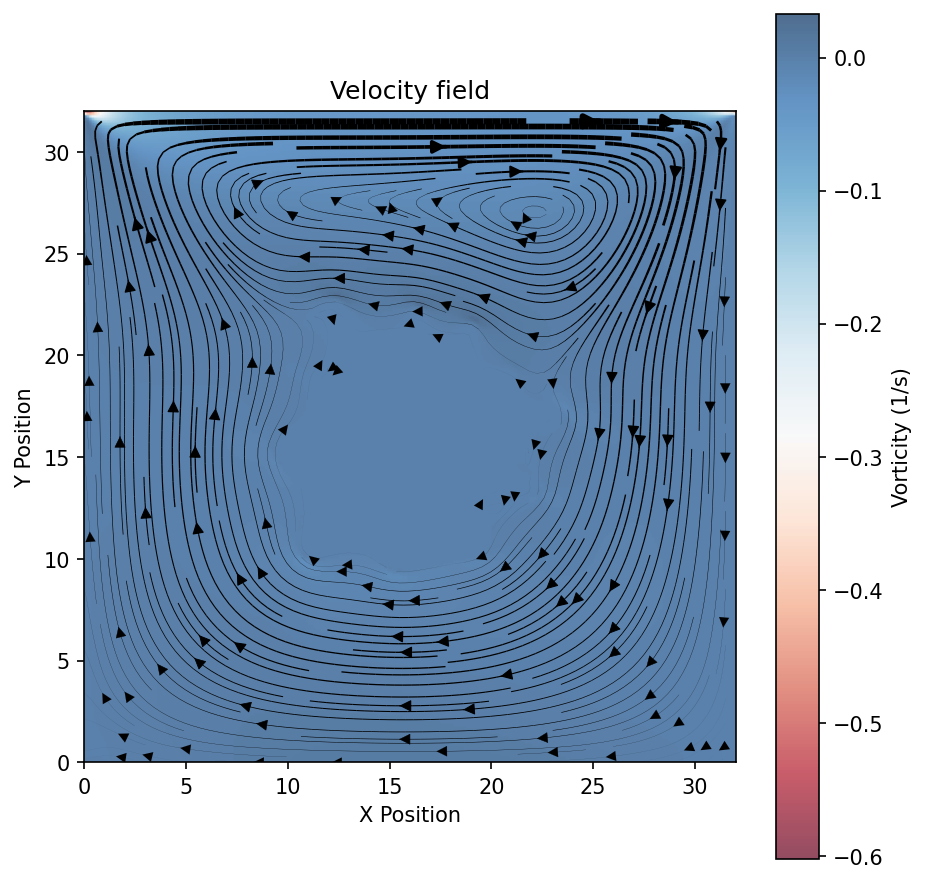}} \\
\multicolumn{2}{c}{\footnotesize Source: FlowBench/LDC\_NS\_2D/256x256/harmonics\_lid\_driven\_cavity\_Y/frame\_376\_vortex.png} \\

\midrule
User &
Judge the type of flow field based on the coupled effect of velocity field and pressure field. \\
FieldLVLM &
Lid-Driven Cavity Flow. Through comprehensive analysis, this flow is identified as a lid-driven cavity flow. Supporting evidence includes: The velocity vector matrix reveals characteristic structures: a strong shear layer at the top and recirculation zones at the bottom; Pressure gradients intensify significantly along the normal direction to the moving wall, aligning with the mixed Couette-Poiseuille flow characteristics; A secondary separation point induced by the primary vortex is detected at coordinate [0.71, 0.59] in the bottom-right corner, with its low-pressure core intensity matching the typical value at Reynolds number Re = 37. \\
\midrule

User &
Judge the type of flow field based on the coupled effect of velocity field and pressure field. \\
LLaVA &
This chart shows a distorted flow field, presenting a very complex shape that includes many distorted curves and points. This type of flow field may be a special type of flow field, such as a twisted flow field or other types of nonlinear flow fields.\\
\midrule

User &
Judge the type of flow field based on the coupled effect of velocity field and pressure field. \\
Llama &
In this chart, we can see the coordinate axis of a velocity field and a curved curve. This curve represents the motion of a fluid in space. Based on the characteristics of the chart, we can determine the type of motion of this fluid:
1. Shape of Curve: The curve presents a curved shape, indicating that the fluid exhibits rotational motion in space. This type of motion is commonly referred to as rotational motion.
2. Curvature of the curve: The curvature of the curve is large, indicating that the fluid is moving at a fast speed in space.
3. The degree of curvature of the curve: The degree of curvature of the curve is significant, indicating that the fluid is moving at a fast speed in space.\\
\midrule

User &
Judge the type of flow field based on the coupled effect of velocity field and pressure field. \\
DeepSeek &
Based on the provided data, which includes position coordinates (X, Y) and vorticity values, the flow field exhibits rotational characteristics due to the presence of non-zero vorticity (ranging from 0.1 to 0.4 1/s).
The flow is rotational and viscous, likely a vortex or shear-driven flow. The pressure field would exhibit radial gradients to balance the rotational motion, but additional pressure data would be needed for precise classification (e.g., Rankine vortex vs. free vortex).\\
\bottomrule
\end{tabularx}
\end{table*}

\begin{table*}[t]
\centering
\caption{Vortex parameter extraction accuracy comparison in cavity flow using vision-language models for structured output generation.}\label{t5}

\begin{tabularx}{\textwidth}{@{}>{\bfseries}l X@{}}
\toprule
\multicolumn{2}{c}{\textbf{Visual input example, Extreme Ironing:}} \\
\midrule
\multicolumn{2}{c}{\includegraphics[width=0.4\linewidth]{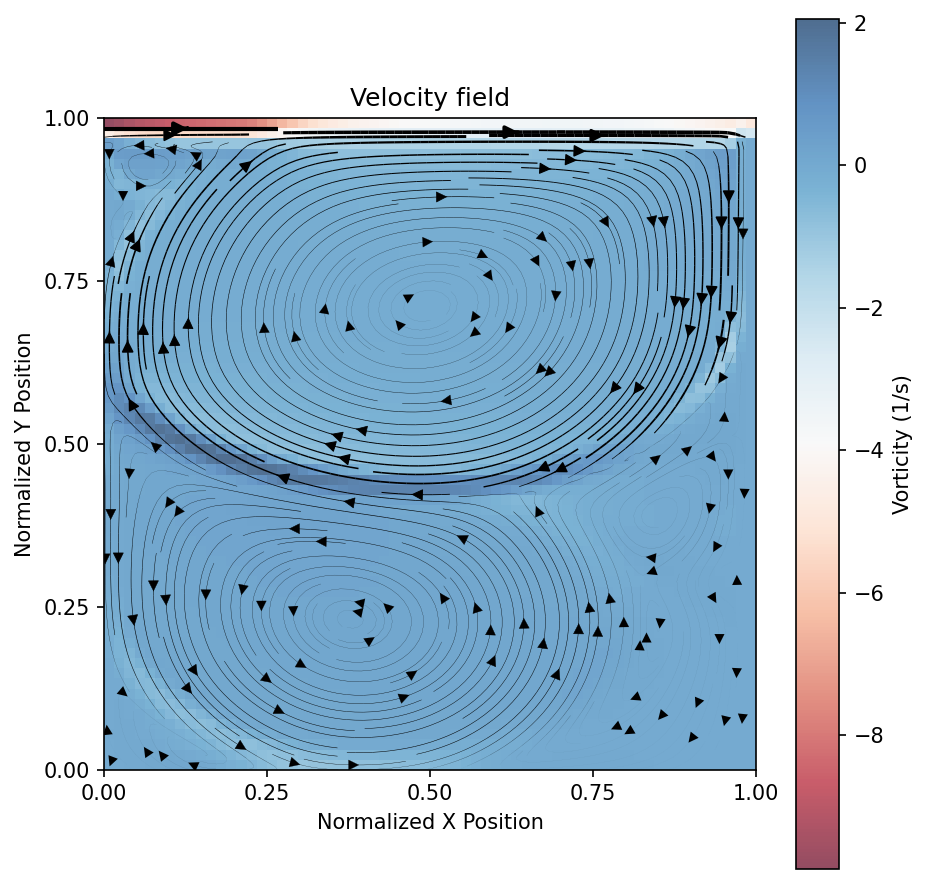}} \\
\multicolumn{2}{c}{\footnotesize Source: CFDBench/cavity/geo/case0011/frame\_18\_vortex.png} \\

\midrule
User &
Please analyze the detailed parameters of all vortex structures.  \\
FieldLVLM &
Two vortices were detected, vortex 1: [length 0.40, height 0.40, circulation 168.36, coordinates (0.38,0.24), rotation direction: counterclockwise] vortex 2: [length 0.45, height 0.45, circulation -142.15, coordinates (0.51,0.72), rotation direction: clockwise] \\
\midrule

User &
Please analyze the detailed parameters of all vortex structures.  \\
LLaVA &
This chart displays the velocity field of a vortex structure, which includes some parameters and distribution patterns.
1. Shape of velocity field: The velocity field of the vortex structure presents a curved curve, which can help us understand the shape of the vortex structure.
2. Distribution of velocity field: The distribution of vortex structures in the velocity field presents a curved shape, which can help us understand the distribution of vortex structures.
3. Strength of velocity field: The vortex structure in the velocity field has a certain strength, which can be measured by the strength value of the velocity field.\\
\midrule

User &
Please analyze the detailed parameters of all vortex structures.  \\
Llama &
This chart shows the parameters and spatial distribution of a vortex structure. This structure is called the 'Velocity field', representing the velocity field. The following is an analysis of the parameters and their spatial distribution patterns in the chart:
1. X Position: This parameter represents the position of the velocity field in the X-axis direction. The chart shows that the velocity field presents a curved curve on the X-axis, indicating that the variation of the velocity field in the X-axis direction is nonlinear.
2. Y Position: This parameter represents the position of the velocity field in the Y-axis direction. The chart shows that the velocity field presents a curved curve on the Y-axis, indicating that the variation of the velocity field in the Y-axis direction is also nonlinear.
3. Velocity: This parameter represents the size of the velocity field. The chart shows that the velocity field presents a curved curve throughout the entire space, indicating that the magnitude of the velocity field also varies nonlinearly in space.\\
\midrule

User &
Please analyze the detailed parameters of all vortex structures.  \\
DeepSeek &
Vortex 1 (Upper Vortex)Location (Core Center):
Vortex 1 (Upper Vortex)Location (Core Center):
Approximate coordinates: (X: 0.50, Y: 0.70)Positioned slightly above the vertical midline. Rotation Direction: Anticyclonic/Counter-Clockwise (inferred from red/yellow vorticity → positive vorticity).Vorticity Magnitude: Peak vorticity: +1.5 to +2.0 $s^{-1}$(centered in red region).Core region: Strong vorticity ($>$ +1.0 $s^{-1}$) extends ~0.15 normalized units radially. Size (Radius): Core radius: ~0.10–0.15(defined by closed streamlines and peak vorticity).
Vortex 2 (Lower Vortex)Location (Core Center):
Approximate coordinates: (X: 0.50, Y: 0.30)Positioned symmetrically below Vortex 1. Rotation Direction: Cyclonic/Clockwise (inferred from blue vorticity → negative vorticity).Vorticity Magnitude: Peak vorticity: -4.0 to -5.0 $s^{-1}$(deep blue core).
Core region: Strong vorticity ($<$ -4.0 $s^{-1}$) extends ~0.10–0.12 normalized units radially. Size (Radius): Core radius: ~0.08–0.12.
\\
\bottomrule
\end{tabularx}
\end{table*}

\textbf{Ablations Study.} Table~\ref{t2} details our component-level investigation. Using Qwen2.5-VL-7B as the baseline (0/NA accuracy), base fine-tuning alone improved performance to 82.28\% through preliminary adaptation to field data characteristics. Subsequent integration of our compression strategy (+Compress data) elevated accuracy to 85.41\%, validating the efficacy of our two-stage pipeline (RGB mapping + VQGAN encoding) in resolving token overflow from excessive data length (e.g., 256×256 matrices). This 99.6\% dimensionality reduction preserved critical physical features while boosting computational efficiency, establishing data compression as essential for resource-intensive scientific applications.
Table~\ref{t3} reveals distinct performance patterns. The same baseline (Qwen2.5-VL-7B) scored 0/NA, while base fine-tuning achieved 53.94\% accuracy – notably lower than other tasks, reflecting inherent complexities of high-dimensional matrices and semantic sparsity. Remarkably, implementing key data selection (+Key data) yielded 100\% accuracy, demonstrating that prioritized analysis of high-value signal regions (e.g., critical flow features) enables efficient global pattern extraction and noise mitigation. This aligns with our core design principle: field data interpretation requires strategic focus on locally significant information.

\begin{table}[t]
\begin{center}
\caption{Quantitative metrics for vortex detection capabilities in field data analysis: performance comparison across vision-language models on core parameters} \label{t6}
\resizebox{0.48\textwidth} {!} {
\begin{threeparttable}
    \begin{tabular}{ c|c|c|c|c }
    \toprule
    Capability &FieldLVLM&LLaVA&Llama&DeepSeek \cr
    \midrule
    \midrule
    Vortex detection count  &\ding{51} (2) &\ding{55} (0)&\ding{55} (0) &\ding{51} (2) \cr
    Core position error &0\% &N/A &N/A &$>$20\% \cr
    Circulation quantification &\ding{51} &\ding{55} &\ding{55} &$\triangle$ (sign error) \cr
    Rotation direction accuracy &\ding{51} &\ding{55} &\ding{55} &\ding{55} (inverted) \cr
    \bottomrule
    \end{tabular}
\end{threeparttable}
}
\end{center}
\end{table}

\subsection{Qualitative results}
Fig.~\ref{fig:5} demonstrates FieldLVLM's responses to four critical queries in a bluff body flow scenario: flow-field classification, vortex structure analysis, Reynolds number estimation, and velocity field interpretation. This test validates the model's ability to extract scientific insights from coupled velocity-pressure field data. FieldLVLM accurately identified the Kármán vortex street dynamics mechanism by recognizing alternating vortex shedding characteristics. For vortex analysis, the model quantified structural parameters of 11 detected vortices (including radii, dimensions, and rotation directions), demonstrating exceptional vortex detection capability. In velocity field analysis, FieldLVLM identified the peak velocity value (1.28 m/s at X=0.92, Y=0.36) and maximum pressure point (X=0.14, Y=0.45), with detailed three-aspect characterization of velocity-pressure variations.

Fig.~\ref{fig:6} examines FieldLVLM's performance on transient dam-break flow through four tasks: velocity distribution analysis, vorticity characterization, flow classification, and Reynolds number estimation. The model accurately located the high-velocity shear layer ([0.23,0.07] to [0.39,0.28]) and identified vorticity concentration zones at the fluid impact front. Classifying the flow as "instantaneous evolution state of dam-break flow," FieldLVLM matched the physical characteristics through velocity gradient analysis while leveraging heatmap visualization features to extract semantic information from RGB mappings.

Table~\ref{t4} compares the performance of FieldLVLM, LLaVA, Llama, and DeepSeek in classifying lid-driven cavity flow based on the coupled effect of velocity and pressure fields. When prompted to "judge the flow type using coupled velocity-pressure field effects", FieldLVLM accurately identifies the flow as lid-driven cavity flow while describing shear layer formation and corner vortex structures, matching classical fluid mechanics principles. In contrast, LLaVA exhibits generalization errors (e.g., labeling it a "distorted flow field") and ignores cavity-specific features. Llama overemphasizes curve geometry (describing "rotational motion") without associating pressure gradients, and DeepSeek provides only partially correct responses (mentioning "shear-driven flow") but lacks quantitative parameters. Crucially, FieldLVLM generates structured responses rich in domain-specific terminology (e.g., "shear layer", "recirculation zones"), exposing other models' knowledge gaps and validating the semantic-structure preservation capability of its RGB mapping + VQGAN compression mechanism.

Table~\ref{t5}  compares the vortex characterization capabilities of FieldLVLM, LLaVA, Llama, and DeepSeek on velocity fields containing coherent structures. When prompted to analyze vortex parameters, FieldLVLM accurately identifies two vortices and reports full physical descriptors. Vortex 1 exhibits counterclockwise rotation with a circulation of +168.36, core coordinates (0.38, 0.24), and size 0.40×0.40. Vortex 2 rotates clockwise with a circulation of -142.15, located at (0.51, 0.72), and measures 0.45×0.45. This demonstrates FieldLVLM's precise integration of position, vorticity, and geometric metrics. In contrast, LLaVA provides vague geometric descriptions (e.g., “curved curves”), failing to capture rotational mechanics. Llama offers non-physical interpretations, treating vector fields as scalar functions and ignoring vortex structures. DeepSeek detects both vortices but introduces critical errors: core positions deviate by over 20\%, circulation magnitudes are overestimated (e.g., -5.0 $s^{-1}$. actual -4.73 $s^{-1}$), and rotation directions are reversed. Only FieldLVLM generates complete, domain-relevant responses using terms like “circulation,” “core coordinates,” and “rotation direction,” validating the effectiveness of its VQGAN-RGB compression in preserving vorticity topology.

\subsection{Qualitative evaluation}
Table~\ref{t6} further quantifies model performance. FieldLVLM achieves perfect accuracy in vortex count, core positioning, circulation magnitude, and rotation direction—highlighting its robust understanding of vortex dynamics. In contrast, DeepSeek mislocates vortex centers, misestimates circulation, and frequently misclassifies rotational direction. These results affirm FieldLVLM's superior capability in extracting and reasoning over structured physical parameters in fluid flow fields.

\section{Conclusion}
This paper proposes a novel method, Improving Large Vision-Language Models’ Understanding for Field Data. Our approach first designs a field-aware language generation strategy which combines the advantages of the high accuracy of special models and the consistency of the large language models, as our data source. Then, the data-compressed multimodal model is presented to compress data for adapting the size of the input of LVLM. It achieves excellent accuracy on field data. In addition, we establish a dedicated evaluation benchmark to assess the capabilities of LVLMs in this domain. While this work focuses on four representative tasks, it serves as an initial step toward broader integration of LVLMs in scientific field analysis. We hope this research encourages further exploration into applying vision-language models to complex scientific data and supports the development of more robust, generalizable multimodal systems for scientific discovery.

\section*{Acknowledgement}
This work was supported by the Strategic Priority Research Program of Chinese Academy of Sciences under Grant XDA0480103, Chinese National Natural Science Foundation Projects 62206280, National Key R\&D Program of China (No. 2025ZD0122000), Chinese National Natural Science Foundation Projects 92570119, Young Scientists Fund of The State Key Laboratory of Multimodal Artificial Intelligence Systems ES2P100113.

\ifCLASSOPTIONcaptionsoff
  \newpage
\fi

\bibliographystyle{IEEEtran}
\bibliography{egbib}

\begin{thebibliography}{10}
\providecommand{\url}[1]{#1}
\csname url@samestyle\endcsname
\providecommand{\newblock}{\relax}
\providecommand{\bibinfo}[2]{#2}
\providecommand{\BIBentrySTDinterwordspacing}{\spaceskip=0pt\relax}
\providecommand{\BIBentryALTinterwordstretchfactor}{4}
\providecommand{\BIBentryALTinterwordspacing}{\spaceskip=\fontdimen2\font plus
\BIBentryALTinterwordstretchfactor\fontdimen3\font minus
  \fontdimen4\font\relax}
\providecommand{\BIBforeignlanguage}[2]{{%
\expandafter\ifx\csname l@#1\endcsname\relax
\typeout{** WARNING: IEEEtran.bst: No hyphenation pattern has been}%
\typeout{** loaded for the language `#1'. Using the pattern for}%
\typeout{** the default language instead.}%
\else
\language=\csname l@#1\endcsname
\fi
#2}}
\providecommand{\BIBdecl}{\relax}
\BIBdecl

\bibitem{radford2021learning}
A.~Radford, J.~W. Kim, C.~Hallacy, A.~Ramesh, G.~Goh, S.~Agarwal, G.~Sastry,
  A.~Askell, P.~Mishkin, J.~Clark \emph{et~al.}, ``Learning transferable visual
  models from natural language supervision,'' in \emph{International conference
  on machine learning}.\hskip 1em plus 0.5em minus 0.4em\relax PmLR, 2021, pp.
  8748--8763.

\bibitem{yuan2021florence}
L.~Yuan, D.~Chen, Y.-L. Chen, N.~Codella, X.~Dai, J.~Gao, H.~Hu, X.~Huang,
  B.~Li, C.~Li \emph{et~al.}, ``Florence: A new foundation model for computer
  vision,'' \emph{arXiv preprint arXiv:2111.11432}, 2021.

\bibitem{li2022grounded}
L.~H. Li, P.~Zhang, H.~Zhang, J.~Yang, C.~Li, Y.~Zhong, L.~Wang, L.~Yuan,
  L.~Zhang, J.-N. Hwang \emph{et~al.}, ``Grounded language-image
  pre-training,'' in \emph{Proceedings of the IEEE/CVF conference on computer
  vision and pattern recognition}, 2022, pp. 10\,965--10\,975.

\bibitem{zhong2022regionclip}
Y.~Zhong, J.~Yang, P.~Zhang, C.~Li, N.~Codella, L.~H. Li, L.~Zhou, X.~Dai,
  L.~Yuan, Y.~Li \emph{et~al.}, ``Regionclip: Region-based language-image
  pretraining,'' in \emph{Proceedings of the IEEE/CVF conference on computer
  vision and pattern recognition}, 2022, pp. 16\,793--16\,803.

\bibitem{zou2023generalized}
X.~Zou, Z.-Y. Dou, J.~Yang, Z.~Gan, L.~Li, C.~Li, X.~Dai, H.~Behl, J.~Wang,
  L.~Yuan \emph{et~al.}, ``Generalized decoding for pixel, image, and
  language,'' in \emph{Proceedings of the IEEE/CVF conference on computer
  vision and pattern recognition}, 2023, pp. 15\,116--15\,127.

\bibitem{li2022language}
B.~Li, K.~Q. Weinberger, S.~Belongie, V.~Koltun, and R.~Ranftl,
  ``Language-driven semantic segmentation,'' \emph{arXiv preprint
  arXiv:2201.03546}, 2022.

\bibitem{bai2025qwen2}
S.~Bai, K.~Chen, X.~Liu, J.~Wang, W.~Ge, S.~Song, K.~Dang, P.~Wang, S.~Wang,
  J.~Tang \emph{et~al.}, ``Qwen2. 5-vl technical report,'' \emph{arXiv preprint
  arXiv:2502.13923}, 2025.

\bibitem{esser2021taming}
P.~Esser, R.~Rombach, and B.~Ommer, ``Taming transformers for high-resolution
  image synthesis,'' in \emph{Proceedings of the IEEE/CVF conference on
  computer vision and pattern recognition}, 2021, pp. 12\,873--12\,883.

\bibitem{devlin2019bert}
J.~Devlin, M.-W. Chang, K.~Lee, and K.~Toutanova, ``Bert: Pre-training of deep
  bidirectional transformers for language understanding,'' in \emph{Proceedings
  of the 2019 conference of the North American chapter of the association for
  computational linguistics: human language technologies, volume 1 (long and
  short papers)}, 2019, pp. 4171--4186.

\bibitem{raffel2020exploring}
C.~Raffel, N.~Shazeer, A.~Roberts, K.~Lee, S.~Narang, M.~Matena, Y.~Zhou,
  W.~Li, and P.~J. Liu, ``Exploring the limits of transfer learning with a
  unified text-to-text transformer,'' \emph{Journal of machine learning
  research}, vol.~21, no. 140, pp. 1--67, 2020.

\bibitem{radford2018improving}
A.~Radford, K.~Narasimhan, T.~Salimans, I.~Sutskever \emph{et~al.}, ``Improving
  language understanding by generative pre-training,'' 2018.

\bibitem{brown2020language}
T.~Brown, B.~Mann, N.~Ryder, M.~Subbiah, J.~D. Kaplan, P.~Dhariwal,
  A.~Neelakantan, P.~Shyam, G.~Sastry, A.~Askell \emph{et~al.}, ``Language
  models are few-shot learners,'' \emph{Advances in neural information
  processing systems}, vol.~33, pp. 1877--1901, 2020.

\bibitem{chowdhery2023palm}
A.~Chowdhery, S.~Narang, J.~Devlin, M.~Bosma, G.~Mishra, A.~Roberts, P.~Barham,
  H.~W. Chung, C.~Sutton, S.~Gehrmann \emph{et~al.}, ``Palm: Scaling language
  modeling with pathways,'' \emph{Journal of Machine Learning Research},
  vol.~24, no. 240, pp. 1--113, 2023.

\bibitem{ouyang2022training}
L.~Ouyang, J.~Wu, X.~Jiang, D.~Almeida, C.~Wainwright, P.~Mishkin, C.~Zhang,
  S.~Agarwal, K.~Slama, A.~Ray \emph{et~al.}, ``Training language models to
  follow instructions with human feedback,'' \emph{Advances in neural
  information processing systems}, vol.~35, pp. 27\,730--27\,744, 2022.

\bibitem{biswas2023role}
S.~S. Biswas, ``Role of chat gpt in public health,'' \emph{Annals of biomedical
  engineering}, vol.~51, no.~5, pp. 868--869, 2023.

\bibitem{jia2021scaling}
C.~Jia, Y.~Yang, Y.~Xia, Y.-T. Chen, Z.~Parekh, H.~Pham, Q.~Le, Y.-H. Sung,
  Z.~Li, and T.~Duerig, ``Scaling up visual and vision-language representation
  learning with noisy text supervision,'' in \emph{International conference on
  machine learning}.\hskip 1em plus 0.5em minus 0.4em\relax PMLR, 2021, pp.
  4904--4916.

\bibitem{liu2023visual}
H.~Liu, C.~Li, Q.~Wu, and Y.~J. Lee, ``Visual instruction tuning,''
  \emph{Advances in neural information processing systems}, vol.~36, pp.
  34\,892--34\,916, 2023.

\bibitem{2023InstructBLIP}
W.~Dai, J.~Li, D.~Li, A.~M.~H. Tiong, J.~Zhao, W.~Wang, B.~Li, P.~Fung, and
  S.~Hoi, ``Instructblip: Towards general-purpose vision-language models with
  instruction tuning,'' 2023.

\bibitem{feigenbaum1977art}
E.~A. Feigenbaum \emph{et~al.}, ``The art of artificial intelligence: Themes
  and case studies of knowledge engineering,'' 1977.

\bibitem{zheng2025large}
Y.~Zheng, H.~Y. Koh, J.~Ju, A.~T. Nguyen, L.~T. May, G.~I. Webb, and S.~Pan,
  ``Large language models for scientific discovery in molecular property
  prediction,'' \emph{Nature Machine Intelligence}, pp. 1--11, 2025.

\bibitem{tshitoyan2019unsupervised}
V.~Tshitoyan, J.~Dagdelen, L.~Weston, A.~Dunn, Z.~Rong, O.~Kononova, K.~A.
  Persson, G.~Ceder, and A.~Jain, ``Unsupervised word embeddings capture latent
  knowledge from materials science literature,'' \emph{Nature}, vol. 571, no.
  7763, pp. 95--98, 2019.

\bibitem{jumper2021highly}
J.~Jumper, R.~Evans, A.~Pritzel, T.~Green, M.~Figurnov, O.~Ronneberger,
  K.~Tunyasuvunakool, R.~Bates, A.~{\v{Z}}{\'\i}dek, A.~Potapenko
  \emph{et~al.}, ``Highly accurate protein structure prediction with
  alphafold,'' \emph{nature}, vol. 596, no. 7873, pp. 583--589, 2021.

\bibitem{fan2023improving}
L.~Fan, D.~Krishnan, P.~Isola, D.~Katabi, and Y.~Tian, ``Improving clip
  training with language rewrites,'' \emph{Advances in Neural Information
  Processing Systems}, vol.~36, pp. 35\,544--35\,575, 2023.

\bibitem{gadre2023datacomp}
S.~Y. Gadre, G.~Ilharco, A.~Fang, J.~Hayase, G.~Smyrnis, T.~Nguyen, R.~Marten,
  M.~Wortsman, D.~Ghosh, J.~Zhang \emph{et~al.}, ``Datacomp: In search of the
  next generation of multimodal datasets,'' \emph{Advances in Neural
  Information Processing Systems}, vol.~36, pp. 27\,092--27\,112, 2023.

\bibitem{lai2023scarcity}
Z.~Lai, H.~Zhang, W.~Wu, H.~Bai, A.~Timofeev, X.~Du, Z.~Gan, J.~Shan, C.-N.
  Chuah, Y.~Yang \emph{et~al.}, ``From scarcity to efficiency: Improving clip
  training via visual-enriched captions,'' 2023.

\bibitem{nguyen2023improving}
T.~Nguyen, S.~Y. Gadre, G.~Ilharco, S.~Oh, and L.~Schmidt, ``Improving
  multimodal datasets with image captioning,'' \emph{Advances in Neural
  Information Processing Systems}, vol.~36, pp. 22\,047--22\,069, 2023.

\bibitem{changpinyo2021conceptual}
S.~Changpinyo, P.~Sharma, N.~Ding, and R.~Soricut, ``Conceptual 12m: Pushing
  web-scale image-text pre-training to recognize long-tail visual concepts,''
  in \emph{Proceedings of the IEEE/CVF conference on computer vision and
  pattern recognition}, 2021, pp. 3558--3568.

\bibitem{schuhmann2022laion}
C.~Schuhmann, R.~Beaumont, R.~Vencu, C.~Gordon, R.~Wightman, M.~Cherti,
  T.~Coombes, A.~Katta, C.~Mullis, M.~Wortsman \emph{et~al.}, ``Laion-5b: An
  open large-scale dataset for training next generation image-text models,''
  \emph{Advances in neural information processing systems}, vol.~35, pp.
  25\,278--25\,294, 2022.

\bibitem{zhu2024deepseek}
Q.~Zhu, D.~Guo, Z.~Shao, D.~Yang, P.~Wang, R.~Xu, Y.~Wu, Y.~Li, H.~Gao, S.~Ma
  \emph{et~al.}, ``Deepseek-coder-v2: Breaking the barrier of closed-source
  models in code intelligence,'' \emph{arXiv preprint arXiv:2406.11931}, 2024.

\bibitem{he2016deep}
K.~He, X.~Zhang, S.~Ren, and J.~Sun, ``Deep residual learning for image
  recognition,'' in \emph{Proceedings of the IEEE conference on computer vision
  and pattern recognition}, 2016, pp. 770--778.

\bibitem{zhang2020bridging}
S.~Zhang, C.~Chi, Y.~Yao, Z.~Lei, and S.~Z. Li, ``Bridging the gap between
  anchor-based and anchor-free detection via adaptive training sample
  selection,'' in \emph{Proceedings of the IEEE/CVF conference on computer
  vision and pattern recognition}, 2020, pp. 9759--9768.

\bibitem{hu2022lora}
E.~J. Hu, Y.~Shen, P.~Wallis, Z.~Allen-Zhu, Y.~Li, S.~Wang, L.~Wang, W.~Chen
  \emph{et~al.}, ``Lora: Low-rank adaptation of large language models.''
  \emph{ICLR}, vol.~1, no.~2, p.~3, 2022.

\bibitem{Llama-3.2}
\BIBentryALTinterwordspacing
Meta, ``Llama,'' 2024. [Online]. Available: \url{https://www.llama.com/}
\BIBentrySTDinterwordspacing

\bibitem{tali2024flowbench}
R.~Tali, A.~Rabeh, C.-H. Yang, M.~Shadkhah, S.~Karki, A.~Upadhyaya,
  S.~Dhakshinamoorthy, M.~Saadati, S.~Sarkar, A.~Krishnamurthy \emph{et~al.},
  ``Flowbench: A large scale benchmark for flow simulation over complex
  geometries,'' \emph{arXiv preprint arXiv:2409.18032}, 2024.

\bibitem{luo2023cfdbench}
Y.~Luo, Y.~Chen, and Z.~Zhang, ``Cfdbench: A large-scale benchmark for machine
  learning methods in fluid dynamics,'' \emph{arXiv preprint arXiv:2310.05963},
  2023.

\end{thebibliography}

\end{document}